\documentclass[letterpaper, 10 pt, conference]{ieeeconf}  

\IEEEoverridecommandlockouts                              

\usepackage{times}
\usepackage{amsmath}
\usepackage{amssymb}
\usepackage{cite}
\usepackage{sdsmmMath}
\usepackage{roboticsMath}
\usepackage{graphicx}
\usepackage{url}

\newcommand{\pff}{SET~}

\title{\LARGE \bf
The \pff Perceptual Factors Framework:\\Towards Assured Perception for Autonomous Systems
}

\author{Troi Williams$^{1}$
\thanks{$^{1}$The NSF Eddie Bernice Johnson INCLUDES initiative, Re-Imagining STEM Equity Utilizing Postdoc Pathways (\#2217329) partially funded this work. The author is in the Department of Computer Science at the University of Maryland, College Park, MD 20742, USA.
        {\tt\small troiw@umd.edu}}%
}

\begin{document}

\maketitle
\thispagestyle{empty}
\pagestyle{empty}

\begin{abstract}

Future autonomous systems promise significant societal benefits, yet their deployment raises concerns about safety and trustworthiness. A key concern is assuring the reliability of robot perception, as perception seeds safe decision-making. Failures in perception are often due to complex yet common environmental factors and can lead to accidents that erode public trust. To address this concern, we introduce the \pff (Self, Environment, and Target) Perceptual Factors Framework. We designed the framework to systematically analyze how factors such as weather, occlusion, or sensor limitations negatively impact perception. To achieve this, the framework employs SET State Trees to categorize \textit{where} such factors originate and SET Factor Trees to model \textit{how} these sources and factors impact perceptual tasks like object detection or pose estimation. Next, we develop Perceptual Factor Models using both trees to quantify the uncertainty for a given task. Our framework aims to promote rigorous safety assurances and cultivate greater public understanding and trust in autonomous systems by offering a transparent and standardized method for identifying, modeling, and communicating perceptual risks.

\end{abstract}


\section{Introduction}

\begin{figure}[t]
    \centering
    \includegraphics[trim={0 0.5cm 0 1cm},clip,width=0.35\linewidth]{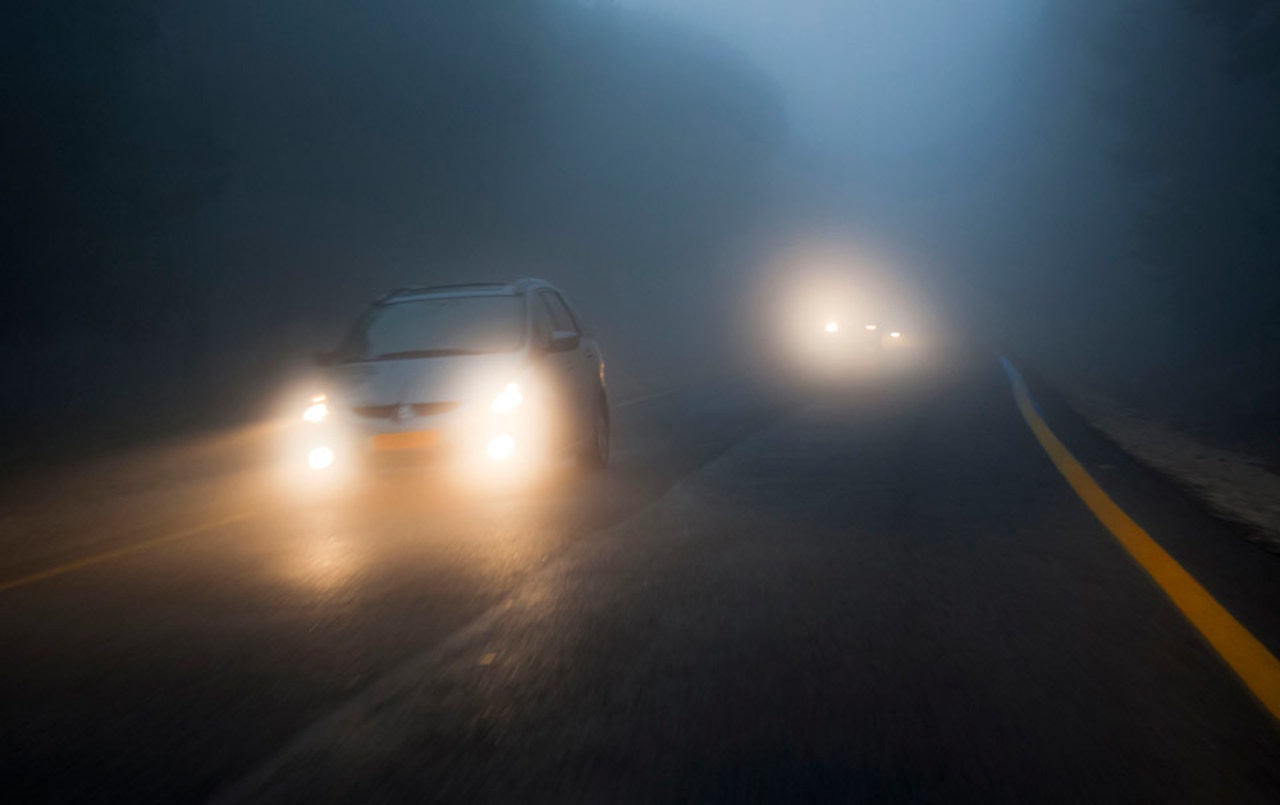}
    \includegraphics[trim={0 0.5cm 0 1cm},clip,width=0.5\linewidth]{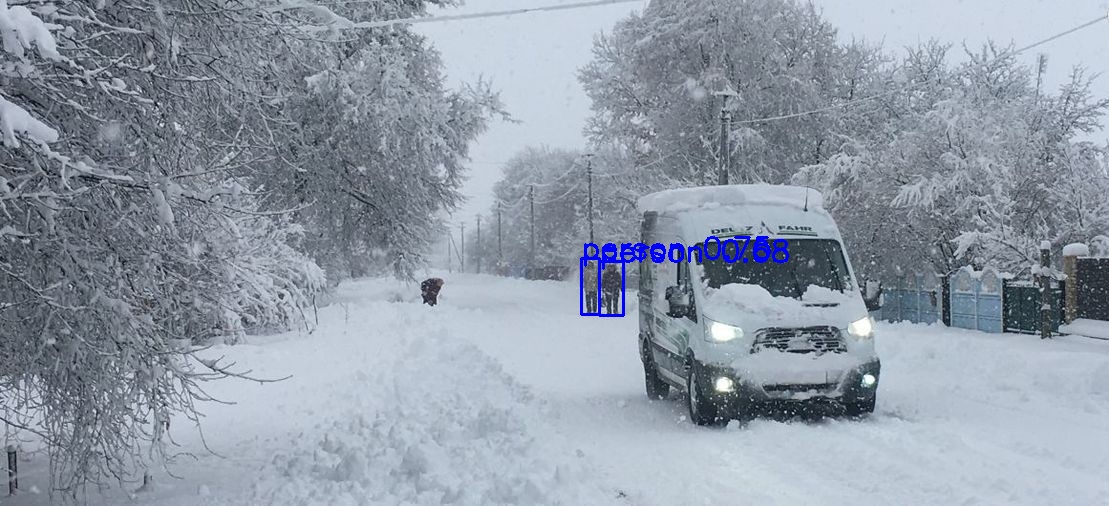}
    \caption{Perceptual failures undermine safety and trust. For example, adverse weather due to haze (left) and snow (right) potentially caused YOLOv8 \cite{Ultralytics_2024} to fail to detect vehicles (images from the DAWN dataset \cite{kenk2020dawn,kenk2020dawndata}). If an autonomous car solely relied on these failed detections, the car could misjudge safety, potentially causing an accident (e.g., during a left turn), injuring occupants, and eroding public confidence.}
    \label{fig:object_detection_with_factors}
\end{figure}

Perceptual tasks like object detection typically seed planning and action \cite{siegwart2011introduction}. Thus, autonomous systems, from autonomous cars to caregiving robots, must perceive their surroundings accurately to operate safely and reliably. However, perception often fails due to factors such as poor visibility, occlusion, or glare (Figure \ref{fig:object_detection_with_factors}). These failures can distort a system's belief of the world and lead to unsafe actions, (severe) accidents, and reduced public trust \cite{Guardian2024tesla,NelkenZitser2023robot,WSJTeslaCrashData2024}.

To mitigate perceptual failures and their consequences, it is crucial to quantify perception uncertainty---how uncertain a system is about its current perceptual predictions. The robotics community has studied perceptual uncertainty extensively (see Section \ref{sec:related} for a brief review) for decades. However, we lack a unifying framework to systematically identify, categorize, analyze, and communicate the factors causing perceptual failures, the sources of these factors, and their impact. Such a gap slows the development of robust safety assurances and makes it difficult for robotists, regulators, and the public to discuss perceptual risks.

We propose the \pff Perceptual Factors Framework to address this gap, where \pff stands for Self (the perceiving agent), Environment, and Target. This framework contributes a structured, human-understandable approach to: 1) identify and model the perceptual factors (like glare) that impact perceptual tasks such as detection or localization; 2) determine how various sources (like the perceiving camera, the sun, and the target car) contribute to these factors; and 3) create Perceptual Factor Models (PFMs) that quantify perceptual uncertainty for given input factors or states.

\section{The \pff Perceptual Factors Framework}

The \pff framework's goal is to model how human-understandable factors and their sources impact perception uncertainty. The framework is composed of three components: a \pff State Tree, a \pff Factor Tree, and a PFM.

\subsection{\pff State Tree: The Detailed Sources of Uncertainty}

\begin{figure}[t]
    \centering
    \includegraphics[trim={0 0.5cm 0 0.5cm},clip,width=0.7\linewidth]{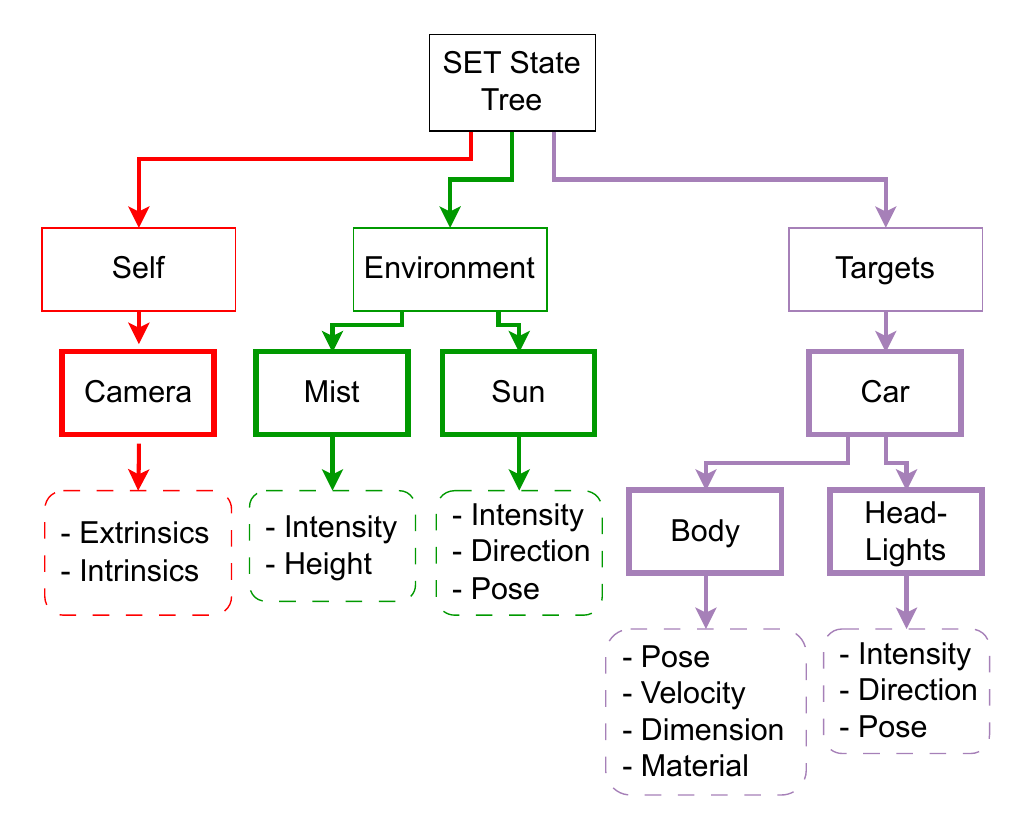}
    \caption{This figure illustrates a sample SET State Tree. The tree models the origins of sources (objects or phenomena) that may impact the car detection task depicted in Figure \ref{fig:object_detection_with_factors} left. This tree also has the same sources depicted in the SET Factor Tree (Figure \ref{fig:set_factor_tree}). Here, ``self'' denotes the perceiving agent's camera, and the targets are the other cars on the road. We use bold boxes for sources and dashed boxes for variables describing the states of sources.}
    \label{fig:state_tree}
\end{figure}

A \pff State Tree \textit{enumerates} the sources (objects or phenomena) that produce perceptual factors and impact perception uncertainty (Figure \ref{fig:state_tree}). This tree helps us and a perceiving agent answer: What attributes of the world and myself (that is, sources and their states) could cause perceptual failures? We organize sources under three branches:
\begin{itemize}
    \item \textbf{Self:} The perceiving agent itself, including its sensors (for example, see ``Self'' in Figure \ref{fig:state_tree}) and other parts of the agent that might affect uncertainty (for example, reflective vehicle surfaces that may cause glare).
    \item \textbf{Environment:} Any object or phenomena, excluding the perceiving agent (self) and the target, such as weather conditions (like rain or fog), light sources (like streetlights), and objects (like trees and buildings).
    \item \textbf{Target:} The objects or phenomena the agent aims to perceive (for example, vehicles in Figure \ref{fig:object_detection_with_factors}).
\end{itemize}

\subsection{\pff Factor Tree: The Chain of Uncertainty}

\begin{figure}[t]
    \centering
    \includegraphics[trim={0 0.5cm 0 0.5cm},clip,width=0.7\linewidth]{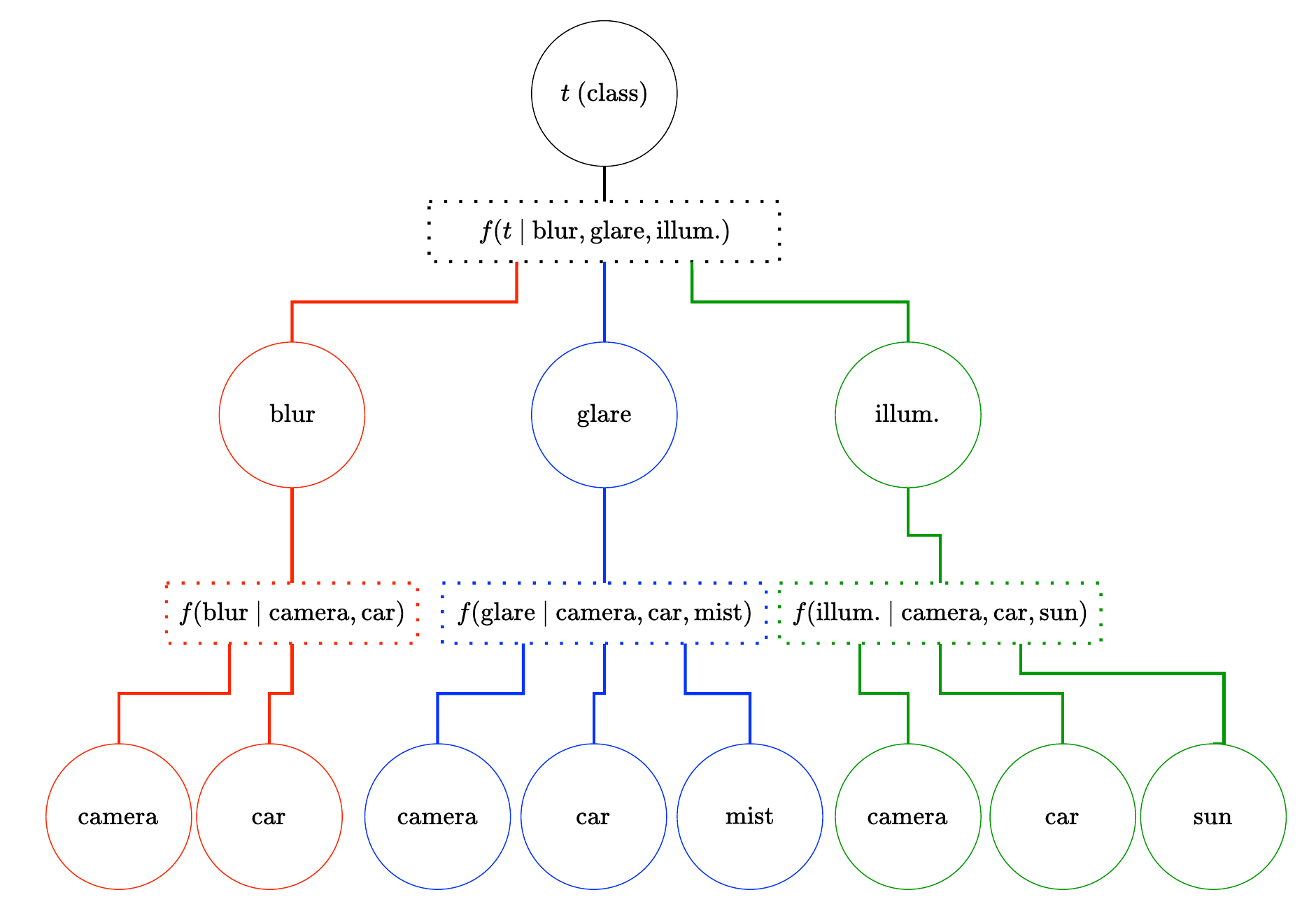}
    \caption{This graph shows a sample SET Factor Tree for the object detection task in Figure \ref{fig:object_detection_with_factors} left. YOLOv8 may have failed to detect the car (target) due to perceptual factors such as motion blur, glare from the headlights, and (low) illumination. These factors could arise due to sources such as the camera, car, mist, and sun. We use dotted boxes to denote factor nodes and circles for variable nodes (observation, perceptual factors, and sources).}
    \label{fig:set_factor_tree}
\end{figure}

A \pff Factor Tree models \textit{how} sources and factors affect perception uncertainty (Figure \ref{fig:set_factor_tree}), showing a chain---from sources to factors and from factors to the perceptual task---to answer two questions. First, how do sources create factors? And second, how do factors impact uncertainty quantitatively (that is, degrade a task)? We can use deterministic or probabilistic methods to model these relationships hierarchically:
\begin{enumerate}
    \item \textbf{Sources:} The leaves of the tree, denoting the objects or phenomena derived from the State Tree.
    \item \textbf{Perceptual Factors:} Intermediate nodes, denoting factors such as glare intensity and motion blur magnitude. These factors arise from interactions between sources. For example, glare is due to sunlight (Environment) shining into a camera (Self), causing image saturation that hinders detection of other vehicles (Target).
    \item \textbf{Perceptual Task:} The root node of the tree, denoting the task we are modeling (e.g., object detection) and whose certainty is affected by perceptual factors.
\end{enumerate}

\subsection{\pff Perceptual Factor Models: Quantifying Uncertainty}

\begin{figure}[t]
    \centering
    \includegraphics[trim={0 10.8cm 0 2cm},clip,width=0.66\linewidth]{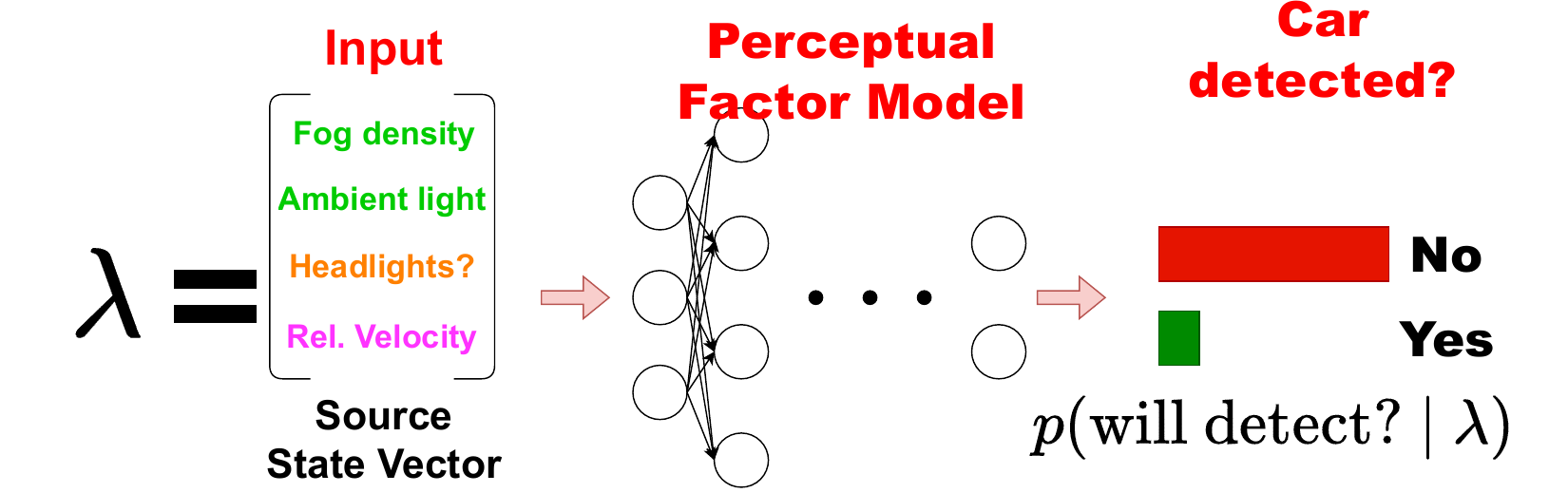}
    \caption{This figure illustrates a sample SET Perceptual Factor Model for a car detection task based on Figure \ref{fig:object_detection_with_factors}. The model predicts if the perceptual algorithm, such as YOLOv8, will detect the car given the source states $\lambda$ \cite{williams2019learning,Sun2020learning,morilla2023perceptual}. The input vector shows examples of source states.}
    \label{fig:pfm}
\end{figure}

We now create a \pff Perceptual Factor Model (PFM), which quantifies perception uncertainty based on the states of the sources (like \cite{williams2019learning,williams2021learningA,Sun2020learning}) or the magnitude of the factors (like \cite{morilla2023perceptual}) (Figure \ref{fig:pfm}). A PFM could be a neural network, Gaussian Process \cite{williams1995GP}, or another probabilistic model. Whether the input of a PFM is states or factors, their values could be estimated, true, or a mixture. For example, a PFM could output the probability of detecting a car for detection tasks or a pose distribution for localization tasks given source states such as camera parameters, atmospheric conditions like fog intensity and ambient lighting, and potential target cars. For either task, we could feed the PFM the camera's state using a Kalman filter onboard the perceiving vehicle, fog intensity using onboard sensors or online weather reports, ambient lighting using onboard ambient light (photodetector) sensors, and \textit{estimated} target (car) states, which hypothesizes the state of the vehicle(s) we are interested in. 

\subsection{Summary: \pff + Public Trust in Autonomous Systems}

This framework promotes trustworthy autonomy in several ways. SET State Trees enable us to systematically identify sources of perceptual failure within various operating environments, reducing overlooked risks compared to ad-hoc testing. The explicit structure (that is, Trees and PFMs) facilitates transparent communication and assurance, allowing us to clearly state considered factors, their modeled impacts, and quantified performance (for example, ``detection probability drops to 90\% in heavy rain'') for safety cases, replacing vague assertions. Furthermore, State Trees and PFMs guide targeted data collection (real or simulated) towards the most challenging and safety-critical scenarios. Finally, combining this framework with others like Failure Modes and Effects Analysis \cite{stamati_1995}, an established safety engineering practice, promotes higher levels of rigorous analysis in perception.

\section{Related Work}\label{sec:related}

\textbf{A Brief Overview:} There is a long history of analyzing how sources and perceptual factors impact perception. Early active vision research \cite{Bajcsy1988activeperception,swain1993promising} explored geometric and optical parameters (camera pose, occlusion, field-of-view) affecting classical object or feature detection methods \cite{cowan1988automatic,cowan1988modelbased,Hutchinson1988planning,Tarabanis1991automatedsensor,Tarabanis1991computingviewpoints,Tarabanis1992occlusionfree,Tarabanis1996occlusionfree,Trucco1997modelbased,Kececi1998improving,Lehel1999sensorplanning,Farag2004image}. Illumination factors like glare \cite{branislav_kisacanin__2006}, back-lighting \cite{Schroeter2009cameraman,lin2006using}, and low light \cite{Tarabanis1991automatedsensor,Tarabanis1995mvp,Tarabanis1995survey} have also been studied extensively. As the popularity of deep learning detectors continues to grow, datasets have been created to mitigate the effects of various visual perceptual factors \cite{ZHANG2023146,appiah2024object,hara2009removalglare,marathe2023wedge,Batchuluun2021deep,chen2021self,Yoneda2021sunglare,Mazhar2021gem,gray2023glaredataset,yahiaoui2020let,esfahani2021robust,alam2024glaremitigation,morawski2021nod,lin2009imagebacklight,hong2022multishiptarget}. Finally, researchers have also modeled how sources and perceptual factors induce heteroscedastic observation noise in self-localization \cite{williams2019learning,Sun2020learning}, active perception \cite{teacy2015observation,morilla2023perceptual,williams2022dynamically}, SLAM \cite{rabiee2021iv}, and detection tasks \cite{Tchuiev2018Inference,Feldman2018bayesian}.

\textbf{How the \pff Framework Differs:} Our framework offers a clear language and structure, via SET Trees, for communicating how perceptual factors affect perception uncertainty and how these factors arise from interactions between \pff sources. This unified approach builds upon related work \cite{williams2019learning,Sun2020learning,morilla2023perceptual} by explicitly modeling both factor impact and origin. It differs from methods focusing primarily on non-interpretable features \cite{rabiee2021iv}, observation correlations \cite{velez2012modelling,teacy2015observation}, specific inference types \cite{Tchuiev2018Inference}, or limited factors \cite{Feldman2018bayesian}.

\addtolength{\textheight}{0cm}   





\bibliographystyle{ieeetr}
\bibliography{root}

\end{document}